\documentclass[10pt]{article}

\usepackage[letterpaper, margin=1in]{geometry}
\usepackage{algorithm}
\usepackage{algorithmic}
\usepackage{caption}
\usepackage{subcaption}
\usepackage{authblk}
\usepackage{setspace}
\usepackage{ourStyle}
\usepackage[inline]{enumitem}
\usepackage{amsmath,amssymb,graphicx,epstopdf,textpos,rotating}

\usepackage[parfill]{parskip}

\newcommand{\expect}{\mathbb{E}}
\allowdisplaybreaks

\usepackage{fancyhdr}

\title{Variance Reduction for Distributed Stochastic Gradient Descent}
\author[S. De et al.]
       {Soham De$^1$, Gavin Taylor$^2$ and Tom Goldstein$^1$\\
       $^1$ Department of Computer Science, University of Maryland, College Park\\
       $^2$ United States Naval Academy \\
       \texttt{\{sohamde, tomg\}@cs.umd.edu, taylor@usna.edu}}

\begin{document}
\maketitle

\begin{abstract}
Variance reduction (VR) methods boost the performance of stochastic gradient descent (SGD) by enabling the use of larger, constant stepsizes and preserving linear convergence rates.   However, current variance reduced SGD methods require either high memory usage or an exact gradient computation (using the entire dataset) at the end of each epoch. This limits the use of VR methods in practical distributed settings. In this paper, we propose a variance reduction method, called VR-lite, that does not require full gradient computations or extra storage. We explore distributed synchronous and asynchronous variants that are scalable and remain stable with low communication frequency. We empirically compare both the sequential and distributed algorithms to state-of-the-art stochastic optimization methods, and find that our proposed algorithms perform favorably to other stochastic methods.
\end{abstract}

\section{Introduction}
Many problems in machine learning and statistics involve minimizations of the
form
\begin{equation}
\label{eq:objective}
\min_x f(x), \hspace{2mm} f(x) = \frac{1}{n}\sum_{i=1}^n f_i(x),
\end{equation}
where each $f_i: \mathbb{R}^d \rightarrow \mathbb{R}.$  Problems of this form include logistic
regression, support vector machines, matrix factorization, and others \cite{recht2011hogwild}.  Such problems frequently arise when fitting a model to data, where each $f_i$ measures how well a model fits a particular data point.  For example, if $\{a_i\}$ is a collection of feature vectors and $\{b_i\}$ is a collection of model outputs, a simple least-squares regression is obtained by setting  $f_i(x) = (a_i^Tx - b_i)^2.$ 
% In the machine learning setting,  $\{b_i\}$ often represents binary labels. In this case, the $f_i$ contain either a logistic link or hinge loss function and the minimization \eqref{eq:objective} is used to train a logistic regression or support-vector machine model.  

% As an example, consider a sequence of $n$ training examples $(a_1, b_1), (a_2, b_2), \cdots, (a_n, b_n)$ where each $a_i \in \mathbb{R}^d$ represents a data sample and $b_i \in \mathbb{R}$ is the label associated with $a_i$. Writing each $f_i(x) = (a_i^Tx - b_i)^2$, we get the standard least regression objective function. Regularization terms can also be included in this setting by including it in each $f_i$. For example, the $L_2$ regularized logistic regression problem for binary classification would have the form $f_i(x) = \ln (1 + \exp(1+b_ia_i^Tx)) + 0.5\lambda\|x\|^2$ where each $b_i \in \{\pm 1\}$.
When large datasets are involved, problems of the form \eqref{eq:objective} are traditionally solved using stochastic gradient descent (SGD) updates of the form
$$x^{k+1} = x^k - \eta^k g^k,$$
where $g^k= \nabla f_{i_k}(x) \approx \nabla
f(x^k)$ is a ``noisy'' gradient estimate using a randomly chosen index $i_k$, and $\eta^k$ is a stepsize
parameter \cite{robbins1951stochastic}. This crude approximate gradient is inexpensive and highly effective
when the error is large, resulting in fast convergence to low accuracy
solutions.  Unfortunately, as the iterates approach the true solution, the
gradient descent stepsize must vanish ($\eta_k\le \mathcal{O}(1/\sqrt k)$) to dampen the
effects of noise, resulting in extremely slow convergence in the high accuracy
regime.

%Some related work here. \cite{de2016efficient, de2016big}

Variance reduction (VR) schemes \cite{johnson2013accelerating, defazio2014saga, reddi2015variance, roux2012stochastic, defazio2014finito, xiao2014proximal, wang2013variance, harikandeh2015stopwasting} are able to maintain a large constant stepsize and achieve fast convergence to high accuracy.    While the difference between the stochastic and exact gradient is potentially large, this difference is predictable -- the gradient errors are highly correlated between different uses of the same function $f_{i_k}.$

VR methods exploit this property by approximating the gradient error at the most recent use of $f_{i_k},$ and subtracting this error from $\nabla f_{i_k}(x^k)$ to obtain a more accurate solution.  This results in approximate gradients of the form 
\begin{align}
g^k=  \underbrace{\nabla f_{i_k}(x)}_{\text{approximate gradient}} -
\underbrace{\nabla f_{i_k}(y) + \widetilde {g_y}}_{\text{error correction
term}}, \label{vr}
\end{align}
where $y$ is an old iterate, and $\widetilde g_y$ is an approximation of the
true gradient $\nabla f(y).$  The algorithm SVRG
\cite{johnson2013accelerating}, for example, sets $y$ to be an iterate from
the algorithm history and sets $\widetilde g_y = \nabla f(y)$ to be the exact
gradient at $y$.  The more recent SAGA scheme \cite{defazio2014saga} stores
the most recent value of $\nabla f_i$ (for all $i$) and chooses  $ \widetilde
g_y$ to be the average of these values.
This error correction term in \eqref{vr} allows VR methods to use large (non-vanishing) stepsizes, preserve linear convergence rates (under strong convexity assumptions), and significantly outperform classical SGD.

Current VR methods suffer from a few caveats that limit their use in practical settings.
First, some methods (e.g., SAGA) require storing a complete history of $n$
previous iterates, thus increasing memory requirements (although
this requirement can be mitigated for some simple minimization problems
\cite{defazio2014saga}).  Other methods (e.g., SVRG) require computation of
the exact gradient $\nabla f$ over the entire dataset after each epoch. This cost is often significant in real-world scenarios where SGD methods typically converge after relatively few epochs, and also prevent its use in asynchronous distributed environments. 

Another caveat of VR algorithms is that they have not been adequately studied in the distributed setting. Current methods are impractical in large scale settings as explained later in the paper, and only a couple of recent papers have tried to address this \cite{de2016efficient, zhao2016scope}.
For large-scale problems, asynchronous variants of distributed SGD are still
widely used \cite{dean2012large, recht2011hogwild, agarwal2011distributed,
li2014communication, shamir2014distributed, zinkevich2009slow,
zinkevich2010parallelized, zhang2015deep}, and most of the work has been focused on the parameter server model of computation
\cite{dean2012large, recht2011hogwild, zinkevich2009slow,
agarwal2011distributed, li2014communication}, where updates are immediately
communicated to the central server. Relatively little work has been done in
the truly distributed case where communications costs are high
\cite{zinkevich2010parallelized, zhang2015deep, mokhtari2016dsa}, since infrequent communication in SGD-type methods usually leads to slower
convergence and instability; a problem that could be mitigated using
VR.

\subsection{Contributions}
This work has two main contributions.  First, we present a new approach to variance reduction, VR-lite,
that requires neither extra memory to store the algorithm history, nor the
computation of exact gradients.  For this reason, VR-lite has lower
memory requirements and faster iterations than previous approaches.  This is
achieved by storing averages of previous iterates rather than a complete
history.  We show empirically that this approach converges much faster than competing methods.

Next, we study the application of VR-lite in a large scale distributed setting. We present synchronous and asynchronous variants of VR-lite in a fully distributed setting with high communication periods between nodes.  We find that our algorithms are scalable and remain stable despite high communication periods. The proposed distributed algorithms outperform a number of other distributed SGD methods on a variety of classification and regression tasks.

\section{Proposed Algorithm: VR-lite}
\label{sec:sequential_algo}

Most VR schemes are divided into epochs, during which a pass is made over the
entire dataset.  Thus, $n$ updates take place during the $m$-th epoch (one
update per data record/feature vector).  The iterates generated in the $m$-th
epoch can be written as $\{x_m^j\}_{j=1}^n.$

At the end of an epoch, the SVRG method sets $y=x_m^n$, and $\widetilde
g_y= \nabla f(y) = \frac{1}{n} \sum_{j=1}^n \nabla f_j(x_m^n)$, the exact gradient.  These
values of $y$ and $\widetilde g_y$  are then used to perform corrected gradient
updates of the form \eqref{vr}.  This approach avoids the extra memory
requirements of ``epoch-free algorithms'' such as SAGA \cite{defazio2014saga}, but
requires an expensive gradient evaluation over the entire dataset on each
iteration.

To speed computation, VR-lite accumulates the {\em average} gradient vector over an epoch.  This average gradient is then used as $\widetilde g_y,$  thus avoiding costly loops over the entire (large) dataset.   
These accumulated averages can be computed cheaply ``on the fly'' as the algorithm runs without noticeable overhead.

Let $\pi_m$ denote a random permutation of the data indices $\{1, 2, \cdots, n\}$, with $\pi_m^j$ denoting the data index chosen in the $j$-th step in $\pi_m$. Then, the update rule for VR-lite is given by
\eqn{update}{
x\kp_{m+1} = x^k_{m+1} - \eta \big(\nabla f_{i_k} (x^k_{m+1}) - \nabla f_{i_k} (\overline{x}_m) + \overline{g}_m \big),
}
where $i_k = \pi_{m+1}^k$, and $\overline{x}_m$ and $\overline{g}_m$ denote the average over iterations and gradients, given by 
\begin{align}
\overline{x}_m = \frac{1}{n}\sum_{j=1}^n x^j_m, \text{ and }
\overline{g}_m = \frac{1}{n} \sum_{j=1}^n \nabla f_{\pi_{m}^j} (x^j_m).
\end{align}
Note that for VR-lite, $\expect[\nabla f_{i_k} (\overline{x}_m) - \overline{g}_m] \neq 0$, unlike most other variance reduction methods. This makes it hard to theoretically prove convergence guarantees for VR-lite.

The values of $\overline{x}_0$ and $\overline{g}_0$ are initialized using a single
epoch of vanilla SGD with no VR correction.  This algorithm fits into the
general VR framework \eqref{vr} with $y =\overline{x}_m$ and $\widetilde g_y =
\overline{g}_m,$ does not require any extra storage apart from a
place to store and update these averages, and only requires 2 gradient
computations per epoch. The full method is described in Algorithm
\ref{alg:sequential}.  

\begin{algorithm}[h]
  \begin{algorithmic}[1]
    \STATE \textbf{parameters} learning rate $\eta$
    \STATE \textbf{initialize} $x$, $\overline{x}$, and $\overline{g}$ using plain SGD for 1 epoch
    \WHILE {not converged}
    \STATE initialize variables to accumulate averages over an epoch: $\widetilde{x} =  \widetilde{g} = 0$
    \FOR {$k$ in $\{1, \dots, n\}$}
    \STATE sample $i_k \in \{1, \dots, n\}$ without replacement
      \STATE update $x$ according to \eqref{update}: $x \gets x - \eta \big(\nabla f_{i_k} (x) - \nabla f_{i_k} (\overline{x}) + \overline{g}\big)$
    \STATE update running averages: $\widetilde{x} \gets \widetilde{x} + x$, $\widetilde{g} \gets \widetilde{g} + \nabla f_{i_j} (x)$
    \ENDFOR
    \STATE set $\overline{x}$ and $\overline{g}$ for next epoch: $\overline{x} = \widetilde{x}/n$, $\overline{g} = \widetilde{g}/n$
%    \STATE set $x$ for next epoch: option 1: $x \gets \overline{x}$; option 2: $x \gets x$
    \ENDWHILE
  \end{algorithmic} 
  \caption{VR-lite Algorithm}  
  \label{alg:sequential}
\end{algorithm}

\section{Distributed Algorithms}
We now consider a setting where the data is distributed across $p$ local
nodes, and our goal is to minimize the global objective function. We consider that the $p$ local nodes can only communicate with a central server, i.e., a centralized setting. Let the set of data indices $\{1,2,\cdots,
n\}$ be decomposed into disjoint subsets $\{\Psi_s\},$ with each $\Psi_s$
representing the indices of data stored on server $s.$   The objective
\eqref{eq:objective} now has the form
\begin{equation}
\label{distributed_obj}
f(x) =  \frac{1}{n} \sum_{s=1}^p  \sum_{j\in \Psi_s} f_j(x). 
\end{equation}

VR-lite is easily distributed in an asynchronous manner since it does not involve calculating the full gradient of the objective function as in SVRG. Moreover, since the algorithm needs to update the average of the gradients only at the end of an epoch, communication periods between the central node and the local nodes can be increased while still ensuring a fast and stable algorithm.

In this section, we propose fast and stable synchronous and asynchronous variants of VR-lite in the fully distributed setting with high communication latency between the central and local nodes.

\subsection{Synchronous VR-lite}
\label{sec:sync_vr}
\begin{algorithm}[h]
  \begin{algorithmic}[1]
    \STATE \textbf{parameters} learning rate $\eta$
    \STATE \textbf{initialize} $x$, $\overline{x}$, $\overline{g}$ using plain SGD for 1 epoch
    \WHILE {not converged}
    \FOR {each local node $s$}
    \STATE initialize variables to accumulate averages over an epoch: $\widetilde{x} =  \widetilde{g} = 0$
    \FOR {$k$ in $\{1, \dots, n\}$}
    \STATE random $i_k \in \Psi_s$ without replacement
      \STATE update $x$ according to \eqref{update}: $x = x - \eta \big(\nabla f_{i_k} (x) - \nabla f_{i_k} (\overline{x}) + \overline{g}\big)$
    \STATE update running averages: $\widetilde{x} = \widetilde{x} + x$, $\widetilde{g} = \widetilde{g} + \nabla f_i (x)$
    \ENDFOR
    \STATE set average variables: $\overline{x}^s = \widetilde{x}/n$, $\overline{g}^s = \widetilde{g}/n$
    \STATE send $x^s$, $\overline{x}^s$, $\overline{g}^s$ to central node
     \STATE receive updated $x$, $\overline{x}$, $\overline{g}$ from central node
    \ENDFOR
    \STATE \textbf{central node:}
    \INDSTATE average all $x^s$, $\overline{x}^s$, $\overline{g}^s$ received from the $p$ local nodes
 %   \INDSTATE set $x$ for next epoch: option 1: $x \gets \overline{x}$; option 2: $x \gets x$
    \INDSTATE broadcast averaged $x$, $\overline{x}$, $\overline{g}$ to each local node $s$
    \ENDWHILE
  \end{algorithmic} 
  \caption{Sync VR-lite Algorithm}  
  \label{alg:sync}
\end{algorithm}
In a fully distributed setting, the objective is to decrease the frequency of (slow)
communication with the central server so as to not impede (fast) gradient
updates on each local node.  This is achieved by updating {\em local} copies
of $x$ on each remote client, and communicating with the central server
periodically to report the change.  Furthermore, the same average gradient
term will be shared across all local nodes and used for variance reduction, ensuring that no
local node drifts far away from the global solution even when communication
gaps are large.  This approach leads to a synchronous algorithm, 
Sync VR-lite, detailed in Algorithm \ref{alg:sync}.

In Sync VR-lite, each local node communicates with the central server only once in each epoch. Thus, during one full epoch over the dataset, Sync VR-lite only requires a total of $p$ communications with the central server. This is much less communication compared to the commonly used ``parameter server" model of computation, where a communication phase is required for each update to a centrally stored iterate, leading to $n$ communications per epoch.

\subsection{Asynchronous VR-lite}
\label{sec:async_vr}

Sync VR-lite can be made asynchronous very easily, as demonstrated in
Algorithm \ref{alg:async}.  The key idea for the asynchronous algorithm, Async VR-lite, is that
only the \emph{change} in $x$, $\overline{x}_s$ and
 $\overline{g}_s$, is sent from local node $s$ to the central
server. This ensures that when updating the centrally stored $x$,
$\overline{x}$, and $\overline{g}$, the previous contribution to the average
from that local worker is subtracted out before adding in the new value. 
Thus, the central $\overline{x}$ and $\overline{g}$ always store
unbiased estimates of the average of the distributed mean $\{\bar x_s\}$ and
$\{\bar g_s\}$
This is critical for ensuring that a fast working local node does not bias the
solution, making the algorithm more robust to heterogeneous computing
environments where local nodes work at drastically different speeds.

\begin{algorithm}[h]
  \begin{algorithmic}[1]
    \STATE \textbf{parameters} learning rate $\eta$
    \STATE \textbf{initialize} $x$, $\overline{x}$ and $\overline{g}$ using plain SGD for 1 epoch; $\alpha = 1/p, x_\text{old} = \overline{x}_\text{old} = \overline{g}_\text{old} = 0$
    \WHILE {not converged}
    \FOR {each local node $s$}
    \STATE initialize variables to accumulate averages over an epoch: $\widetilde{x} =  \widetilde{g} = 0$
    \FOR {$k$ in $\{1, \dots, n\}$}
    \STATE random $i_k \in \Psi_s$ without replacement
      \STATE update $x$ according to \eqref{update}:  $x = x - \eta \big(\nabla f_i (x) - \nabla f_i (\overline{x}) + \overline{g}\big)$
    \STATE update running averages: $\widetilde{x} = \widetilde{x} + x$, $\widetilde{g} = \widetilde{g} + \nabla f_i (x)$
    \ENDFOR
    \STATE set average variables: $\overline{x} = \widetilde{x}/n$, $\overline{g} = \widetilde{g}/n$
    \STATE calculate the change in variables: $\Delta{x}^s = x - x_\text{old}$, $\Delta{\overline{x}}^s = \overline{x} - \overline{x}_\text{old}$, $\Delta{\overline{g}}^s = \overline{g} - \overline{g}_\text{old}$
    \STATE saving the current average variables: $\overline{x}_\text{old} = \overline{x}$, $\overline{g}_\text{old} = \overline{g}$
    \STATE send $\Delta{x}^s$, $\Delta{\overline{x}}^s$, $\Delta{\overline{g}}^s$ to central node
     \STATE receive updated $x$, $\overline{x}$, $\overline{g}$ from central node
    \ENDFOR
    \STATE \textbf{central node:}
    \INDSTATE receive $\Delta{x}^s$, $\Delta{\overline{x}}^s$, $\Delta{\overline{g}}^s$ from a local worker $s$
    \INDSTATE update central variables: $x = x + \alpha \Delta{x}$, $\overline{x} = \overline{x} + \alpha \Delta{\overline{x}}$, $\overline{g} = \overline{g} + \alpha \Delta{\overline{g}}$
 %   \INDSTATE set $x$ for next epoch: option 1: $x \gets \overline{x}$; option 2: $x \gets x$
    \INDSTATE send new $x$, $\overline{x}$, $\overline{g}$ back to local worker $s$
    \ENDWHILE
 \end{algorithmic} 
  \caption{Async VR-lite Algorithm}  
  \label{alg:async}
\end{algorithm}

\section{Experiments}
\label{sec:results}

\begin{figure*}[t]
  \centering
  \begin{subfigure}[t]{0.24\textwidth}
    \includegraphics[width=\textwidth]{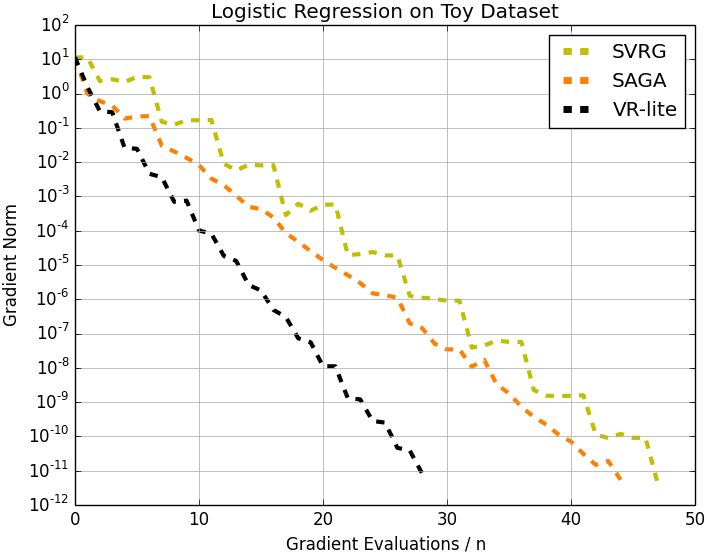}
    %\caption{Logistic regression on toy data set}
    \label{fig:log_toy}
  \end{subfigure}
  \hfill
  \begin{subfigure}[t]{0.24\textwidth}
    \includegraphics[width=\textwidth]{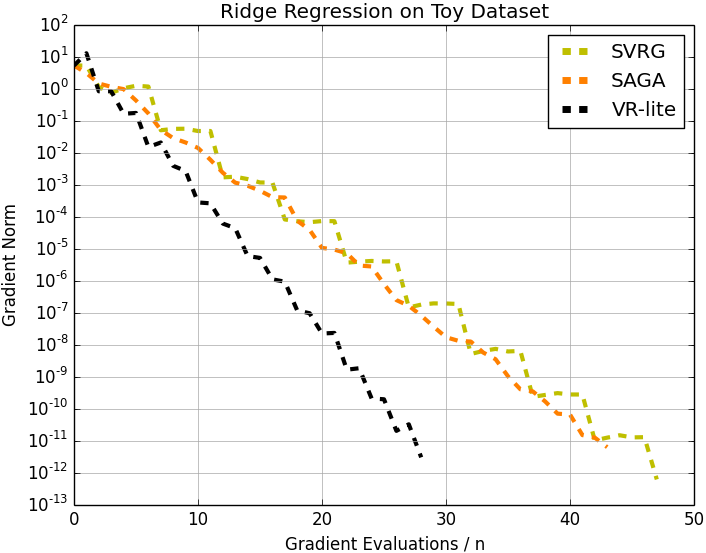}
    %\caption{Ridge regression on toy data set}
    \label{fig:lin_toy}
  \end{subfigure}
  \hfill
  \begin{subfigure}[t]{0.24\textwidth}
    \includegraphics[width=\textwidth]{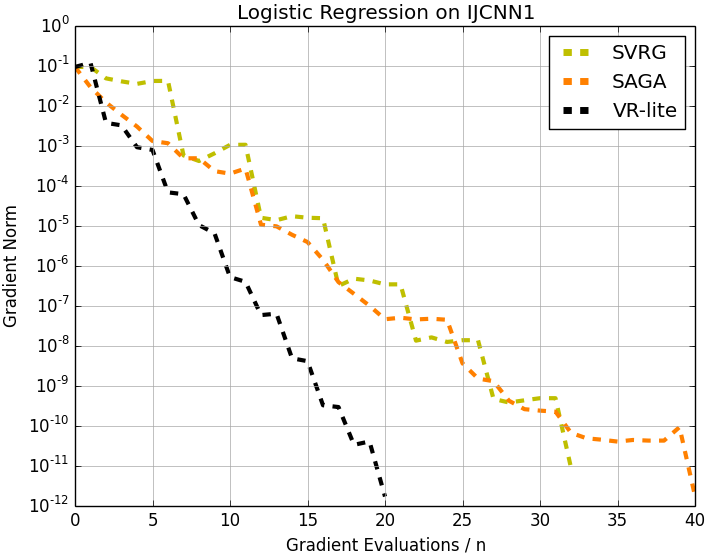}
    %\caption{Logistic Regression on IJCNN}
    \label{fig:ijcnn}
  \end{subfigure}
  \hfill
  \begin{subfigure}[t]{0.24\textwidth}
    \includegraphics[width=\textwidth]{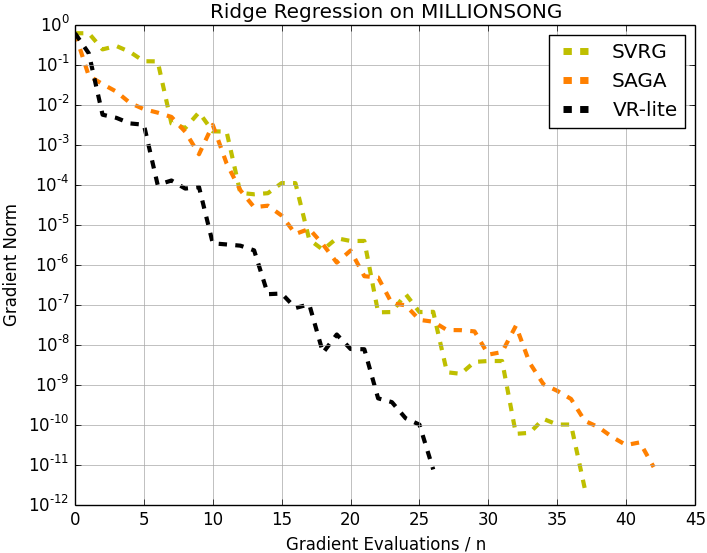}
    %\caption{Logistic Regression on MILLIONSONG}
    \label{fig:ijcnn}
  \end{subfigure}
    \vspace{-6mm} 
  \caption{\small Sequential Results. Left to right: Logistic regression on toy dataset; Ridge regression on toy data; Logistic regression on IJCNN1 dataset; Ridge regression on MILLIONSONG dataset}
  \label{fig:seq}
\end{figure*}

\begin{figure*}[t]
  \centering
  \begin{subfigure}[t]{0.24\textwidth}
    \includegraphics[width=\textwidth]{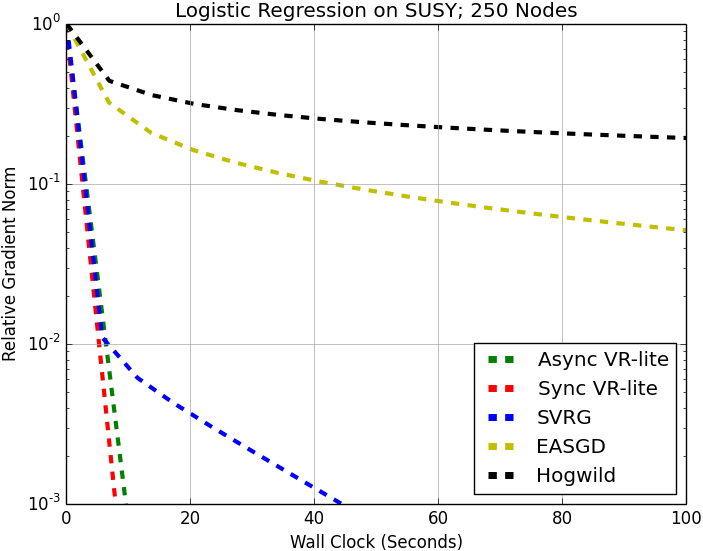}
    %\caption{Logistic regression on toy data set}
    \label{fig:log_toy}
  \end{subfigure}
  \hfill
  \begin{subfigure}[t]{0.24\textwidth}
    \includegraphics[width=\textwidth]{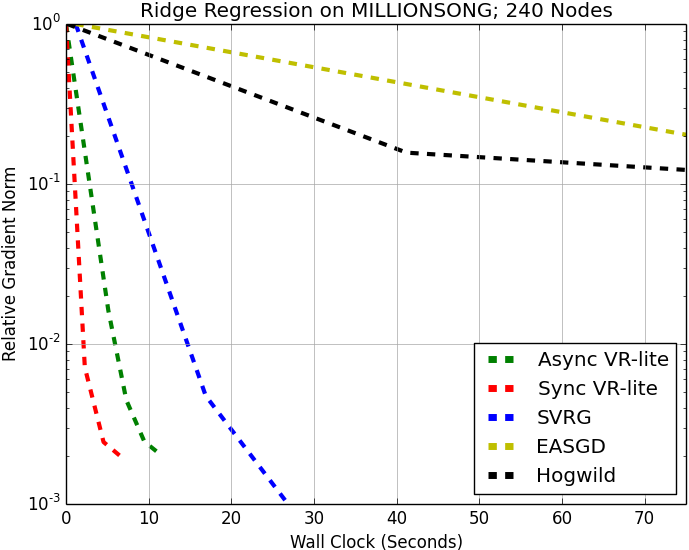}
    %\caption{Ridge regression on toy data set}
    \label{fig:lin_toy}
  \end{subfigure}
  \hfill
  \begin{subfigure}[t]{0.24\textwidth}
    \includegraphics[width=\textwidth]{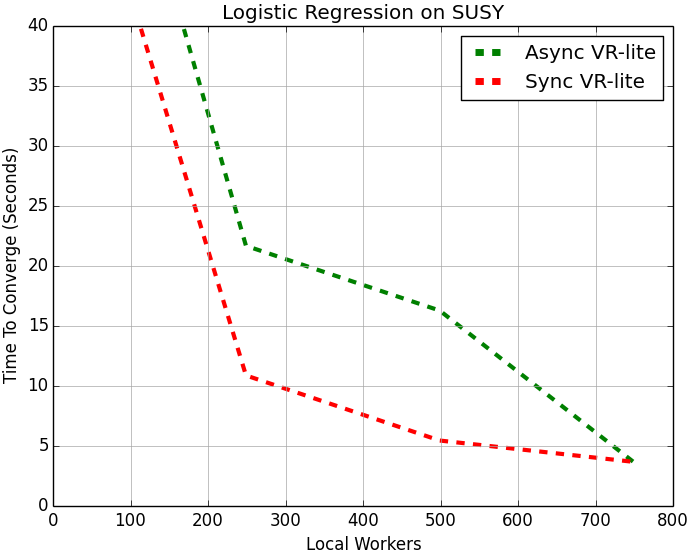}
    %\caption{Logistic Regression on IJCNN}
    \label{fig:ijcnn}
  \end{subfigure}
  \hfill
  \begin{subfigure}[t]{0.24\textwidth}
    \includegraphics[width=\textwidth]{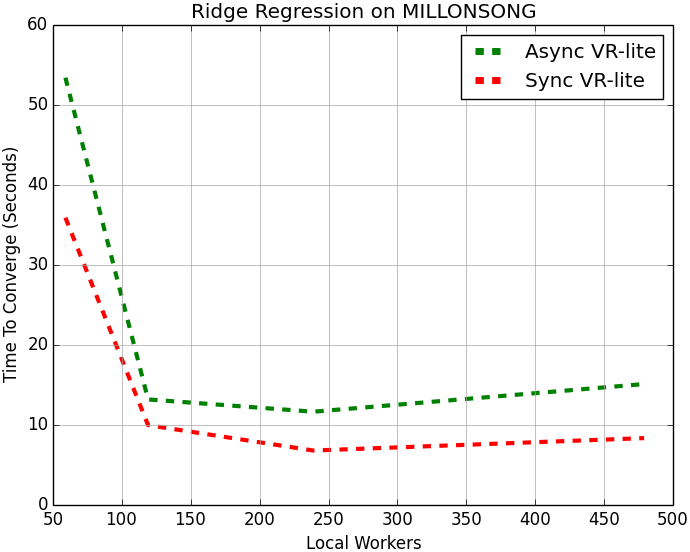}
    %\caption{Logistic Regression on MILLIONSONG}
    \label{fig:ijcnn}
  \end{subfigure}
    \vspace{-6mm} 
  \caption{\small Distributed Results. Left to right: Logistic regression on SUSY over 250 local workers; Ridge regression on MILLIONSONG over 240 local workers; Time required for convergence as number of local workers is increased on SUSY and MILLIONSONG, respectively.}
  \label{fig:dist}
\end{figure*}

In this section, we present our empirical results. We test our algorithms on a
binary classification problem with $\ell_2$-regularized logistic regression and an
$\ell_2$-regularized linear regression (ridge regression) problem. More
formally, we are interested in the following two optimization problems:
\begin{align}
\text{logistic regression:  }\quad & \min_x \frac{1}{n} \sum_{i=1}^n \log \big(1+ \exp(b_ia_i^Tx) \big) + \lambda   \|x\|^2  \nonumber \\
 \text{ridge regression:  }\quad  & \min_x \frac{1}{n} \sum_{i=1}^n (a_i^Tx-b_i)^2 + \lambda \|x\|^2, \nonumber
\end{align}
where each $a_i \in \mathbb{R}^d$ is a data sample with label $b_i \in \mathbb{R}$, and $x \in \mathbb{R}^d$. In all our experiments, we set the $\ell_2$ regularization parameter to $\lambda = 10^{-4}$. 

\subsection{Sequential Results}
We first investigate the performance of VR-lite in the sequential, non-distributed setting. It is well-established by now that VR methods out-perform regular SGD by a wide margin on convex models. Thus, we compare our algorithm's performance with two popular variance reduction methods, SVRG \cite{johnson2013accelerating} and SAGA \cite{defazio2014saga}.

We tested VR-lite on two toy datasets as well as two real world datasets. For binary classification, we generated a random Gaussian data matrix for each class, where the first class had zero mean and the second class had mean 1 entries. The variances of the distributions were varied such that the dataset was not perfectly linearly separable. For ridge regression, we generated a random Gaussian matrix $A$ with observation vector $b = Ax + \epsilon$, where $\epsilon$ denotes standard Gaussian noise. We built the toy datasets to have $n = 5000$ data samples with $d = 20$ features, with 2500 points of each class for the binary classification case. We also tested our method on two standard real world datasets: IJCNN1 \cite{prokhorov2001ijcnn} for binary classification and the MILLIONSONG \cite{Bertin-Mahieux2011} dataset for least squares regression.

Figure \ref{fig:seq} shows results of our experiments. For all algorithms, we pick the constant learning rate which gives fastest convergence.
Since a gradient computation is the most expensive step of VR methods, we compare the rate of convergence with the number of epochs used. It is important to note that comparing VR-lite to SAGA may not be a fair comparison since SAGA uses considerably more storage space. Despite this fact, we find that VR-lite converges faster than both SAGA and SVRG in all cases.
 
\subsection{Distributed Results}
\label{dist_results}
We now present results of our distributed algorithms. All
algorithms were implemented using Python and MPI, and run on an HPC cluster
with 24 cores per node. Our asynchronous implementations are ``locked", i.e., only a single local node can update the parameters on the central server at a given time. However, all our implementations can be made faster in a simple extension to a lock-free setting.  We compare our algorithms to the following popular
asynchronous stochastic optimization methods:
\begin{itemize}
\item Hogwild!: A basic asynchronous variant of SGD \cite{recht2011hogwild} using a ``parameter server'' model.
\item Elastic Averaging SGD (EASGD): A recent asynchronous
  SGD method \cite{zhang2015deep} that has been shown to accelerate the
  training of neural networks efficiently. We found the basic
  EASGD algorithm performed better than the momentum method (M-EASGD), so we
  present results only on EASGD. As recommended in \cite{zhang2015deep}, we
  set the parameter $\beta = 0.9$. The communication period was varied as $\tau = {1, 4, 16, 64}$.
\item Asynchronous SVRG: This is the first work to our knowledge to propose an asynchronous variant of VR algorithms \cite{reddi2015variance}. The authors show that the asynchronous SVRG algorithm enjoys the same benefits as its sequential counterpart over standard distributed stochastic optimization methods. As recommended in both \cite{johnson2013accelerating} and \cite{reddi2015variance}, we set the epoch size to $2n$. Note that, after each epoch of size $2n$, a synchronization step is required for this algorithm to calculate the full gradient. Thus, this is not a fully asynchronous method.
\end{itemize}

Note that there is no accepted standard stochastic optimization algorithm for the distributed setting, and so we choose to compare VR-lite with the above recently developed algorithms.

For all algorithms and all experiments, we present results using the constant step size that achieves fastest convergence. 
We present results for binary classification on the dataset SUSY \cite{baldi2014searching} and for least squares regression on the dataset MILLIONSONG \cite{Bertin-Mahieux2011}. MILLIONSONG contains over 500,000 data samples while SUSY contains 5,000,000 samples.

Figure \ref{fig:dist} shows results of our distributed experiments. The first two plots compare our algorithms with the other methods on SUSY with 250 local workers and MILLIONSONG with 240 local workers. We plot the relative norm of the gradient on the $y$-axis, and wall clock time on the $x$-axis. For both the binary classification and least squares regression problems, we see that both Sync VR-lite and Async VR-lite significantly outperform all other algorithms. We further notice that the methods are extremely stable inspite of the high communication latency with the central server.

The last two plots in Figure \ref{fig:dist} show the scalability of our algorithms. We plot the wall clock time required for convergence on the $y$-axis and increase the number of local workers on the $x$-axis. We compare convergence rates of Sync VR-lite and Async VR-lite with 60, 120, 240 and 480 local workers for MILLIONSONG and with 50, 250, 500, 750 local workers for SUSY. We notice that for MILLIONSONG, there is a large drop in convergence rate initially with the gains leveling out as we keep increasing the number of local workers. The proposed algorithms take only about 10 seconds over hundreds of local workers to train the dataset and the convergence rates of the asynchronous algorithm can potentially be further improved by using a lock-free implementation. For the dataset SUSY, which is 10 times larger than MILLIONSONG, we see a consistent gain in convergence rates as we keep increasing the number of local workers, with our algorithms training the dataset in less than 5 seconds when distributed over 750 local workers.

\section{Conclusion}
In this paper, we presented VR-lite, a variance reduction SGD algorithm which, unlike other popular variance reduction schemes, does not require any full gradient computations or extra storage. We presented distributed variants of this algorithm, Sync VR-lite and Async VR-lite, for a setting with low frequency of communication between the local nodes and the central server. Sync VR-lite and Async VR-lite are scalable in the distributed setting, and perform favorably to other parallel and distributed SGD algorithms on standard models typically encountered in machine learning.

{\small
\bibliography{references}
\bibliographystyle{plain}
}

\end{document}